\documentclass[runningheads]{llncs}
\usepackage{caption}
\usepackage{booktabs}
\usepackage{graphicx}
\usepackage{adjustbox}
\usepackage{amssymb}
\usepackage{ulem}
\usepackage{amsmath}
\usepackage{bigfoot}
\usepackage{etoolbox}
\usepackage{marvosym}
\usepackage[hidelinks]{hyperref}

\begin{document}
\title{RetiGen: A Framework for Generalized Retinal Diagnosis Using Multi-View Fundus Images}

\titlerunning{Multi-View Domain Generalization for Retinal Diagnosis}

\author{Ze Chen\inst{3}\textsuperscript{(*)} \and
Gongyu Zhang\inst{1}\textsuperscript{,}\inst{2}\textsuperscript{(*,\Letter)} \and
Jiayu Huo\inst{1} \and
Joan Nunez do Rio\inst{1} \and
Charalampos Komninos\inst{1} \and
Yang Liu\inst{1} \and
Rachel Sparks\inst{1} \and
Sebastien Ourselin\inst{1} \and
Christos Bergeles\inst{1} \and
Timothy Jackson\inst{2}}

\authorrunning{Z. Chen et al.}
\institute{School of Biomedical Engineering \& Imaging Sciences, King's College London \and
School of Life Course \& Population Sciences, King's College London \and
University of Electronic Science and Technology of China 
\\
}

\maketitle



\begingroup
\renewcommand\thefootnote{*}
\footnotetext{These authors contributed equally to this work.}
\renewcommand\thefootnote{\Letter}
\footnotetext{Corresponding author: Gongyu Zhang, email: gongyu.zhang@kcl.ac.uk.}
\endgroup

\begin{abstract}
This study introduces a novel framework for enhancing domain generalization in medical imaging, specifically focusing on utilizing unlabelled multi-view colour fundus photographs. Unlike traditional approaches that rely on single-view imaging data and face challenges in generalizing across diverse clinical settings, our method leverages the rich information in the unlabelled multi-view imaging data to improve model robustness and accuracy. By incorporating a class balancing method, a test-time adaptation technique and a multi-view optimization strategy, we address the critical issue of domain shift that often hampers the performance of machine learning models in real-world applications. Experiments comparing various state-of-the-art domain generalization and test-time optimization methodologies show that our approach consistently outperforms when combined with existing baseline and state-of-the-art methods. We also show our online method improves all existing techniques. Our framework demonstrates improvements in domain generalization capabilities and offers a practical solution for real-world deployment by facilitating online adaptation to new, unseen datasets. Our code is available at https://github.com/zgy600/RetiGen.
\keywords{Domain generalization  \and Multi-view \and Diabetic Retinopathy}
\end{abstract}
\section{Introduction}

A variety of deep learning models are developed for accurate image analysis and pathology diagnosis reaching high performance in their specific domains and image modalities, e.g. optical coherence tomography (OCT)~\cite{Zhang2021-Clinically}, magnetic resonance imaging~\cite{mazurowski2019deep}, computed tomography (CT)~\cite{primakov2022automated}, and ultrasound~\cite{liu2019deep}. However, common challenges in medical imaging, such as multiple acquisition techniques and devices, variability in image quality or patient anatomy, and the subtle differences in pathology manifestation, often lead to the use of previously trained models in new data or domains. 

In that context, the need for techniques to address domain gap adaption is crucial. Existing state-of-the-art approaches use exclusively single-view imaging data and traditional machine learning or deep learning models. For example, Fishr~\cite{rame2022fishr} encourages the pursuit of flatter minima during the training process, $\mathrm{D}^{3}\mathrm{G}$\cite{yao2023improving} leverages domain similarities based on domain metadata to learn domain-specific models, and MAT\cite{wang2022improving} employs adversarial training by combining various perturbations at the domain level. 
Most existing methodologies are optimized for performance on known datasets and often fail to adapt seamlessly to unseen domains. This inability to adapt in the absence of labelled data from the new domain restricts the practical deployment of models, especially in healthcare settings where collecting labelled data can be challenging due to privacy concerns and the need for expert annotation.
Moreover, these methods generally do not support the integration of multi-view imaging data from the target domain, if available, which could potentially offer richer and more diverse information for model training and adaptation.

\begin{figure}[tb]
    \centering
    \includegraphics[width=\columnwidth]
    {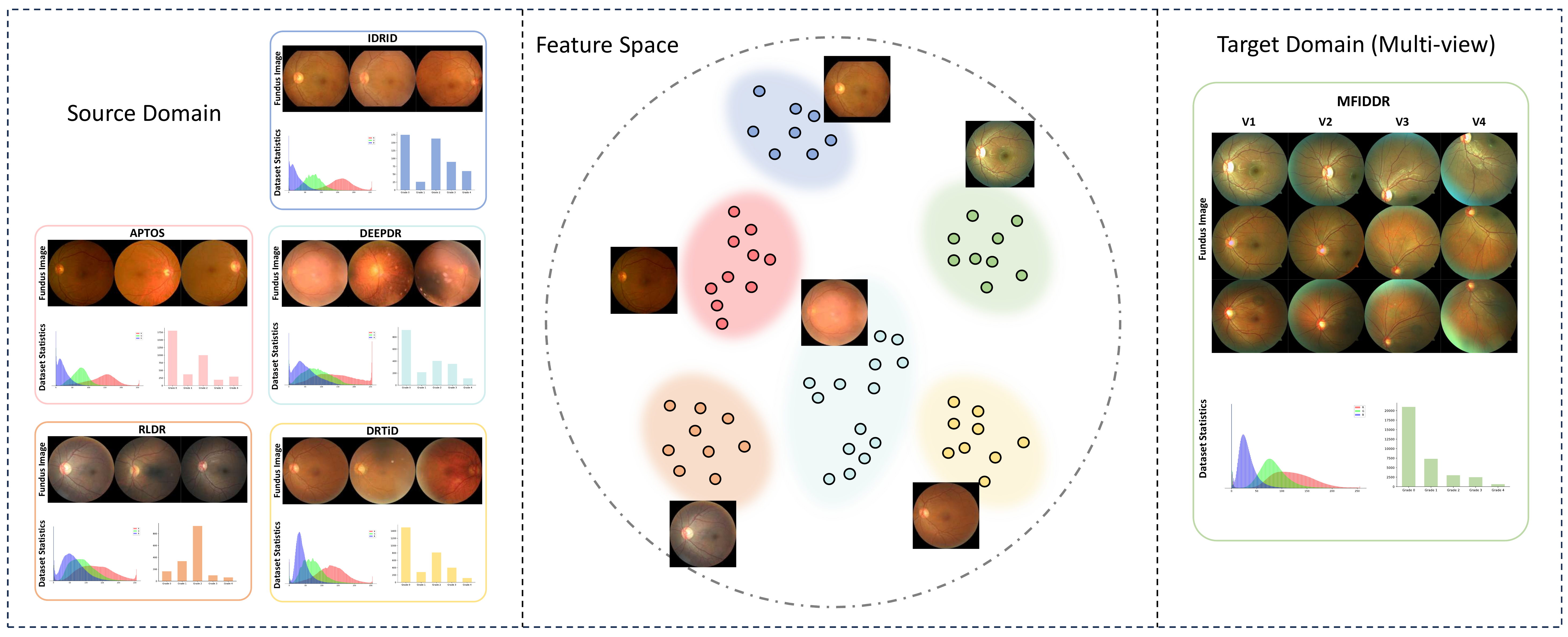}
    \caption{The figure presents our experimental setup, emphasizing the utilization of multi-view images in the target domain, specifically incorporating four distinct views for each patient: the field centred on the macula (\(V_1\)), the field centred on the optic disc (\(V_2\)), and the fields tangent to the upper and lower horizontal lines of the optic disc, respectively (\(V_3\) and \(V_4\)).}

    \label{fig:multi-view-dgdr}
\end{figure}

Multi-view learning has gained relevance in recent years~\cite{yan2021deep,jin2023deep}. Multi-view images are prevalent in various medical imaging modalities, such as colour fundus photography, mammography, and computed tomography, capturing different angles and perspectives of the subject~\cite{luo2023MVCINN,li2020multi}. While existing methodologies have been developed to process multi-view images~\cite{wang2023swinmm,LI2023455,karthik2021deep,shachor2020mixture,zeng2021multi},
they often require both training and inferencing strictly on multi-view data, limiting their application in real-world scenarios. In practice, most deep learning models are trained and tested on single-view imaging domains, despite the availability of unlabelled multi-view images in some target domains. Procedures to achieve data adaptation to a multi-view target domain (Fig.\ \ref{fig:multi-view-dgdr}) without the need for model architecture alteration or re-optimization with labelled data can facilitate a model's potential uses.

In this study, we present a Framework for \textit{Generalized Retinal Diagnosis Using Multi-View Fundus Images} (RetiGen), drawing inspiration from domain generalization and test-time optimization studies~\cite{chokuwa2023generalizing,wang2021tent}. RetiGen utilizes multi-view images to advance domain generalization in the test time, which is applicable in both online and offline scenarios. Our approach encompasses a three-step process designed to address key challenges in domain generalization: \textit{Pseudo-label based Distribution Calibration} (PDC) addresses class imbalance through pseudo-labels to recalibrate data distribution; \textit{Test-time Self-Distillation with Regularization} (TSD) refines decision boundaries using test-time data to improve model robustness; and \textit{Multi-view Local Clustering and Ensembling} (MVLCE) exploits multi-view images to enrich the feature space, enhancing pseudo-label accuracy and decision-making through k-nearest neighbour searches. 

This work contributes to the field in several ways. First, we establish a novel application scenario within domain generalization, specifically focusing on the utilization of unlabelled multi-view medical imaging data from the target domain. Second, we develop a unique framework that leverages multi-view imagery to enhance domain generalization in medical imaging, incorporating techniques such as multi-view decision boundary refinement and test-time adaptation to address the challenges of domain shift. Third, we conduct comprehensive evaluations, integrating our framework with existing state-of-the-art domain generalization methods, and further benchmark its performance against leading test-time optimization techniques. Our results demonstrate notable improvements across all methods, underscoring the effectiveness of our approach in enhancing model robustness and accuracy in diverse clinical settings.

\section{Methodology}

\begin{figure}[tb]
    \centering
    \includegraphics[width=\columnwidth]
    {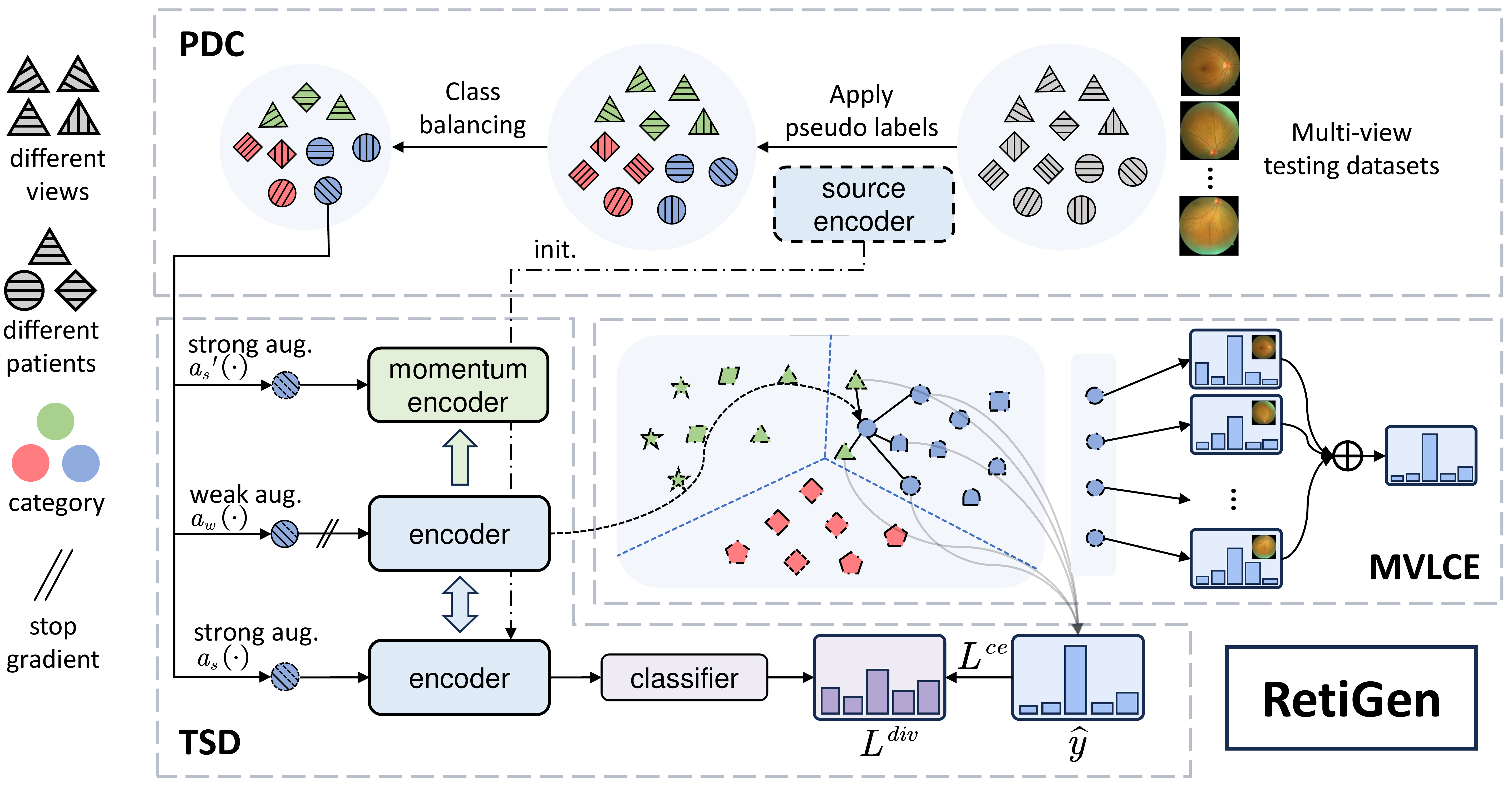}
    \caption{Overview of the proposed domain-generalization framework, beginning with a pseudo-label-based selection from a multi-view target dataset at the top (PDC). Mini-batch images undergo augmentation before encoding in feature space for test-time self-distillation (TSD), enhancing the source model. The process concludes with ensembling refined multi-view image embeddings (MVLCE) of the same patient, integrating diverse perspectives for improved diagnosis accuracy.
    }

    \label{fig:my_label}

\end{figure}
The methodology begins by leveraging a model previously trained on source domains $\mathcal{D}_s$, characterized by parameters $\theta_s \in \Theta_s$ and guided by a loss function $\ell$. Our goal revolves around the application of this source model to unlabeled multi-view target data ${x_i^{(v)}} \in \mathcal{D}_t$, where $i \in {1, \ldots, N}$ indexes the images, $v \in {1, \ldots, V}$ identifies the view within the target domain $\mathcal{D}_t$ and $N$ represents the total number of target images. This strategy is visually represented in Fig.\ \ref{fig:my_label}.

\subsection{Pseudo-label Based Distribution Calibration (PDC)}
To tackle decision bias and overfitting caused by imbalanced datasets—a prevalent issue in medical imaging—we employ Pseudo-label Distribution Calibration (PDC). This technique commences with initialising the target model $g = f \circ h$ using source model parameters $\theta_s$ from the domain $\mathcal{D}_s$. Here, $f$ symbolizes the backbone of the model and $h$ acts as its linear projection head. Upon receiving an unlabeled target sample $x_i$, the model $g$ processes it to extract features $z_i = f(x_i)$, compute logits $p_i = h(z_i)$, and generate pseudo-labels $\hat{y}_i = \arg\max p_i$. To correct imbalances within the dataset, random undersampling is conducted based on these pseudo-labels, adjusting the dataset to a balanced size of $CN_m$, where $N_m$ is the minimal count of initial pseudo-labels for each class and $C$ denotes the total class count. This calibration ensures a fair representation across classes in the target dataset, effectively reducing biases and improving overall model performance. The pseudo-label balanced target domain dataset generated by the PDC will be used to test-time self-distillation Sec. \ref{subsection 2.2}.

\subsection{Test-time Self-Distillation with regularization (TSD)}\label{subsection 2.2}

\subsubsection{Weak-strong consistency and self-distillation}

Once the classes are balanced, we proceed with test-time adaptation, employing strategies to ensure the model's predictions remain consistent across varying degrees of image augmentations. 
Specifically, for a given target image $x_i$, we generate three augmentations: two strong ($a_s(x_i)$, $a_s'(x_i)$) from the same augmentation distribution $\mathcal{A}_s$, and one weak ($a_w(x_i)$) from $\mathcal{A}_w$. This process, as visualized in Fig.\ \ref{fig:my_label}, forms the basis for our adaptation strategy, which does not rely on ground truth labels from the target domain. Instead, it uses pseudo-labels derived from multi-view local clustering to supervise the predictions of strongly augmented samples. 
The encoder is an initialised trained source domain model, providing a robust foundation for subsequent adaptation. 
$\hat{y}^c$ signifies the $c$-th element of the one-hot pseudo-label that undergoes label-smoothing to mitigate overconfidence in the model's predictions, and $p_q=\sigma(g(a_s(x_i)))$ represents the predicted probabilities for the strongly-augmented query images, with $c$ marking the class index. The equation for cross-entropy loss is articulated as follows:
\begin{equation}
L_{t}^{ce}=-\mathbb{E}_{x_i\in\mathcal{D}_t}\sum_{c=1}^C{\hat{y}^c\log p_{q}^{c}},
\end{equation}
\subsubsection{Memory queue}
We employ nearest neighbour search, facilitated by a memory queue, to refine pseudo-label accuracy by comparing newly encountered target samples with previously observed samples stored in the queue. This comparison is based on feature similarity, enabling the model to assign more accurate pseudo-labels by leveraging the collective knowledge encapsulated in the queue. The queue, implemented as a dictionary, stores the feature embeddings and prediction probabilities of weakly augmented target data, with its size decoupled from the mini-batch size to allow for a comprehensive nearest neighbour search database. While using a queue enables the dictionary to expand, it also complicates updating the encoder through backpropagation. To address this challenge and ensure consistency and stability in the feature space, a momentum model $g^{\prime} = f^{\prime} \circ h^{\prime}$ is employed to calculate update features $w'$ and probabilities $p'$. Also, the momentum model is updated at each mini-batch step, initialized with the source encoder weights $\theta_s$:

\begin{equation} 
w'=f'(a_s'(x_i)),\quad p'=\sigma(h'(w'))
\end{equation}

\begin{equation}
\theta_t^{\prime}\leftarrow m\theta_t^{\prime}+(1-m)\theta_t,
\end{equation}
where $\theta_t$ and $\theta_t^{\prime}$ represent the weights of the encoder and momentum encoder, respectively. The use of a momentum coefficient $m$, set to a high value of $0.999$, ensures gradual updates, promoting consistency and reliability in the feature space throughout the domain adaptation process.

\subsubsection{Diversity regularization}
Although pseudo-label refinement can effectively reduce the noise of pseudo-labels caused by domain drift, the potential harm of some erroneous pseudo-labels cannot be ignored. We employ a regularization term to encourage the model not to overly rely on misleading pseudo-labels during the adaptation process during the adaptation process, thereby promoting class diversity. $c$ and $C$ are the index and the number of classes:
\begin{equation} 
L_t^{div}=\mathbb{E}_{x_i\in\mathcal{D}_t}\sum_{c=1}^C\bar{p}_q^c\log\bar{p}_q^c
\end{equation}
\begin{equation} 
\bar{p}_q=\mathbb{E}_{x_i\in\mathcal{D}_t}\sigma(g(a_s(x_i)))
\end{equation}
\subsubsection{The overall loss}
The loss function consists of the weak-strong consistency regularization and the diversity regularization.

\begin{equation} 
L_t=L_t^{ce}+L_t^{div}
\end{equation}

\subsection{Multi-View Local Clustering and Ensembling (MVLCE)}
\subsubsection{Multi-view Local Clustering} 

Due to the gaps between domains associated with variations between different data sources or environments, classifiers may make erroneous decisions for some target samples. However, a more accurate estimation can be obtained by aggregating knowledge from points near the sample feature space of the test dataset $\mathcal{D}_t$ with $\mathcal{M}{\text{ views}}$, thus yielding more realistic pseudo-labels for adaptation. 

We utilize the dense information encapsulated within the feature space of test dataset $\mathcal{D}_t$ with $\mathcal{M}{\text{ views}}$. We compute the cosine distance between the feature vector of the current weakly augmented image and the feature vectors stored in the queue $Q$ to identify the $K$ nearest neighbours most similar to the current image feature. We perform soft voting by ensembling the probabilities of these $K$ neighbours:
\begin{equation}
\hat{p}^{(i,c)}=\frac1K\sum_{j\in\mathcal{I}i}p^{\prime{(j,c)}}
\end{equation}
where $p^{\prime(j,c)}=\sigma(h^{\prime}(w^{\prime(j)}))$ and $\mathcal{I}{i}$ represents the index of the neighbor in the memory queue $Q$ that is closest to the current image feature. Subsequently, we obtain estimates with lower noise, upon which we base our decision for pseudo-labeling:
\begin{equation}
\hat{y}^i=\arg\max\hat{p}^{(i,c)}
\end{equation}

\subsubsection{Multi-view Ensembling} 

Leveraging multi-view fundus images---for example, views centred on the macula (\(V_1\)), optic disc (\(V_2\)), and fields tangent to the optic disc's upper and lower horizontal lines (\(V_3\) and \(V_4\))---enhances diagnostic accuracy by integrating diverse anatomical insights. Our multi-view ensembling method aggregates predictions from these distinct views using soft voting:

\begin{equation}
\mathring{p}^{(i,c)}=\frac{1}{\mathcal{M}}\sum_{j\in \mathcal{V} i}{p^{\prime(j,c)}}
\end{equation}

Here, \(\mathcal{M}\) denotes the total number of views, and \(\mathcal{V}i\) includes different views for the current image. This approach ensures a comprehensive analysis by synthesizing information across multiple perspectives, thereby improving the robustness of the diagnostic outcome.

\section{Results and Discussion}
\noindent \textbf{Experimental settings, implementation details, and evaluation metrics. }
Our choice for the backbone architecture is ResNet50, pre-trained on ImageNet, with a fully connected layer serving as the linear classifier. We reported on three key metrics: accuracy (ACC), area under the ROC curve (AUC), and Macro F1 score (F1). Bold and underline formatting denote the first and second highest scores, respectively. 
We choose five publicly available Diabetic Retinopathy datasets in our experimental setup. Five datasets for training. In the result, Table 1-4, DeepDR~\cite{liu2022deepdrid}, IDRID~\cite{porwal2018indian}, APTOS~\cite{karthick2019aptos}, DRTID~\cite{hou2022cross}, and RLDR~\cite{wei2021learn} are combined to train the source model and a multi-view imaging dataset MFIDDR~\cite{luo2023MVCINN} as target domain dataset for testing. The results of the one-train-one-test setup are also attached in Supplementary Table 1. 

\begin{table}[t]
\caption{Integration of our method with various domain generalization methods. Five different public datasets are combined to train the source dataset. Performance is tested on the MFIDDR multi-view dataset.
}
\label{tab:transposed}
\centering
\resizebox{\textwidth}{!}{%
\begin{tabular}{l|cc|cc|cc|cc|cc|cc|cc}
\toprule
 & ERM~\cite{vapnik1999overview} & w/ Ours & MixStyle~\cite{zhou2021domain} & w/ Ours & Fishr~\cite{rame2022fishr} & w/ Ours & GREEN~\cite{liu2020green} & w/ Ours & Mixup~\cite{zhang2017mixup} & w/ Ours & CABNet~\cite{he2020cabnet} & w/ Ours & GDRNet~\cite{che2023generalizable} & w/ Ours \\
\midrule
AUC & 75.5 & \textbf{81.5} & 69.1 & \textbf{73.2} & 78.0 & \textbf{78.2} & 78.8 & \textbf{82.5} & 79.4 & \textbf{85.8} & 77.7 & \textbf{86.3} & 83.0 & \textbf{87.4} \\
ACC & 53.5 & \textbf{54.5} & 33.0 & \textbf{50.3} & \textbf{60.3} & 55.6 & \textbf{63.8} & 49.9 & 60.5 & \textbf{61.6} & 56.2 & \textbf{60.9} & 51.8 & \textbf{61.5} \\
F1 & 39.4 & \textbf{45.5} & 24.2 & \textbf{36.1} & 41.2 & \textbf{42.8} & 41.7 & \textbf{47.3} & 44.5 & \textbf{53.3} & 43.4 & \textbf{54.6} & 50.4 & \textbf{58.8} \\
\bottomrule
\end{tabular}%
}
\label{tab:comprehensive_evaluation}
\end{table}
\begin{table}[t]
\centering 
\caption{(Left) Comparison with test-time adaptation methods.}
\caption{(Right) Ablation studies.}
\label{tab:comparison_tta} 
\label{tab:ablation} 

\begin{minipage}[b]{.5\textwidth}
\centering
\scriptsize 
\begin{tabular}{cc|ccc}
\toprule
Source & Methods & AUC & ACC & F1 \\
\midrule
 & TENT~\cite{wang2021tent} & 66.8 & 41.6 & 28.4\\
ERM~\cite{vapnik1999overview} & AdaContrast~\cite{chen2022contrastive} & \uline{69.5} & \uline{50.3} & \uline{31.3}\\
 & Ours & \textbf{81.5} & \textbf{54.5} & \textbf{45.5}\\
\midrule
 & TENT~\cite{wang2021tent} & 73.7 & 47.7 & 36.2 \\
GDRNet~\cite{che2023generalizable} & AdaContrast~\cite{chen2022contrastive} & \uline{77.8} & \uline{61.5} & \uline{42.4}\\
 & Ours & \textbf{87.4} & \textbf{61.5} & \textbf{58.8}\\
\bottomrule
\end{tabular}

\end{minipage}
\begin{minipage}[b]{.5\textwidth}
\centering
\small
\begin{tabular}{cc|ccc}
\toprule
TSD & MVLCE & AUC & ACC & F1 \\
\midrule
 & & 83.0 & 51.8 & 50.4\\
\checkmark & & 83.8 & 53.3 & 51.5 \\
 & \checkmark & \uline{86.5} & \uline{58.8} & \uline{55.9} \\
\midrule
\checkmark & \checkmark & \textbf{87.4} & \textbf{61.5} & \textbf{58.8} \\
\bottomrule
\end{tabular}
\label{tab:Ablation}
\end{minipage}
\end{table}


\noindent \textbf{Adding RetiGen domain generalization to the non-test-time adaptation baselines} We conducted a comprehensive set of experiments to evaluate our framework, comparing it with a vanilla baseline (ERM~\cite{vapnik1999overview}) and other state-of-the-art domain generalization methods from different categories in Table \ref{tab:comprehensive_evaluation}, including Mixstyle~\cite{zhou2021domain}, Fishr~\cite{rame2022fishr}, GREEN~\cite{liu2020green}, Mixup~\cite{zhang2017mixup}, CABNet~\cite{he2020cabnet}, GDRNet~\cite{che2023generalizable}. The overall results in Table \ref{tab:comprehensive_evaluation} indicate that our approach can easily integrate with other state-of-the-art techniques and improve performance. 
Within our experimental framework, the proposed approach combined with GDRNet~\cite{che2023generalizable} achieves the best performance, with an increase of 4.4\% AUC and with the final AUC of 87.4\% and an 8.6\% increase for the source model trained on CABNet~\cite{he2020cabnet} to 86.3\%. We meticulously evaluated each source model, where each was trained on distinct source datasets, but uniformly tested on the same target dataset, MFIDDR. This comprehensive analysis, detailed in Supplementary Table 1, consistently reveals that our method significantly boosts performance across a wide array of models and their respective source datasets.

\noindent \textbf{Comparison of test-time adaptation approaches} 
Our method demonstrates superior performance over existing test-time adaptation techniques across all assessed metrics, see Table 2. Specifically, compared to the best-performing alternative, our approach achieves a notable improvement in AUC and F1 score, along with an increase in accuracy. These results underscore the efficacy and superiority of our proposed method in enhancing model performance during test-time adaptation.

\noindent \textbf{Ablation study}
In our ablation study, we evaluate the distinct contributions of TSD and MVLCE components, using a source model trained with GDRNet~\cite{che2023generalizable} and evaluated on the multi-view imaging dataset MFIDDR~\cite{luo2023MVCINN}. Results depicted in Table \ref{tab:ablation} highlight the efficacy of each component. Implementing TSD alone provides a modest enhancement in performance, showcasing its role in fine-tuning predictions through pseudo-label optimization and regularization techniques. The integration of MVLCE leads to significant improvements across all evaluated metrics (elevating AUC from 83.0\% to 86.5\%, ACC from 51.8\% to 58.8\%, and F1 from 50.4\% to 55.9\%), emphasizing the advantage of leveraging multi-view data to enrich the feature space. Merging TSD and MVLCE offers the most substantial performance upgrades. This approach culminates in an AUC of 87.4\%, ACC of 61.5\%, and F1 of 58.8\%, demonstrating the combined power of TSD and MVLCE in enhancing model adaptability and accuracy.

\begin{table}[t]
\caption{The performance of our method (online) integrated with various domain generalization methods is evaluated on the multi-view domain generalization benchmark (ResNet-50 backbone).}
\label{tab:transposed_online}
\centering
\resizebox{\textwidth}{!}{%
\begin{tabular}{l|cc|cc|cc|cc|cc|cc|cc}
\toprule
 & ERM~\cite{vapnik1999overview} & w/ Ours & MixStyle~\cite{zhou2021domain} & w/ Ours & Fishr~\cite{rame2022fishr} & w/ Ours & GREEN~\cite{liu2020green} & w/ Ours & Mixup~\cite{zhang2017mixup} & w/ Ours & CABNet~\cite{he2020cabnet} & w/ Ours & GDRNet~\cite{che2023generalizable} & w/ Ours \\
\midrule
AUC & 75.5 & \textbf{80.7} & 69.1 & \textbf{73.6} & 78.0 & \textbf{82.5} & 78.8 & \textbf{82.1} & 79.4 & \textbf{82.1} & 77.7 & \textbf{82.1} & 83.0 & \textbf{86.5} \\
ACC & 53.5 & \textbf{61.7} & 33.0 & \textbf{33.5} & 60.3 & \textbf{66.7} & 63.8 & \textbf{65.1} & 60.5 & \textbf{65.1} & 56.2 & \textbf{66.8} & 51.8 & \textbf{58.8} \\
F1 & 39.4 & \textbf{42.0} & \textbf{24.2} & 23.8 & 41.2 & \textbf{44.8} & 41.7 & \textbf{46.3} & 44.5 & \textbf{46.3} & \textbf{43.4} & 40.8 & 50.4 & \textbf{56.0} \\
\bottomrule
\end{tabular}%
}
\end{table}

\noindent \textbf{Applying RetiGen in an Online Setting}
The results of the online version of our framework are in Table \ref{tab:transposed_online}, focusing exclusively on Multi-view Local Clustering and Ensembling (MVLCE), showcases an enhanced adaptability feature, setting it apart from its offline counterpart. This distinction arises because, in the online setting, we cannot access the entire distribution of the test set, necessitating a streamlined approach that relies solely on MVLCE. Despite this constraint, the online method still significantly improves key performance metrics such as AUC, ACC, and F1 scores across various domain generalization methods, underscoring its efficacy in real-world application scenarios.





\section{Conclusion}

RetiGen introduces a novel, innovative framework enhancing domain generalization in medical imaging through the strategic use of multi-view fundus photographs. Our comprehensive methodology, incorporating Pseudo-label Distribution Calibration (PDC), Test-time Self-Distillation with Regularization (TSD), and Multi-view Local Clustering and Ensembling (MVLCE), not only addresses the prevalent issue of domain shift but also sets a new benchmark for effectively leveraging multi-view imaging data. PDC recalibrates the data distribution for improved model robustness against class imbalance; TSD refines model accuracy by precisely aligning the model's confidence with its predictions in a new domain; and MVLCE exploits the richness of multi-view data to enhance decision-making through enriched feature spaces. Our comprehensive evaluation demonstrates that integrating RetiGen with any existing domain generalization method yields significant improvements across all performance metrics, emphasizing our approach's universal applicability and effectiveness. This underscores the potential of our method to serve as a powerful, versatile tool for enhancing the accuracy and reliability of diagnostic models in various clinical environments.

\bibliographystyle{splncs04}
\bibliography{mybibliography}

\end{document}